\documentclass[conference]{IEEEtran}
\IEEEoverridecommandlockouts
\usepackage{cite}
\usepackage{amsmath,amssymb,amsfonts}
\usepackage{algorithmic}
\usepackage{graphicx}
\usepackage{textcomp}
\usepackage{xcolor}
\usepackage{booktabs}
\usepackage{float}
\def\BibTeX{{\rm B\kern-.05em{\sc i\kern-.025em b}\kern-.08em
    T\kern-.1667em\lower.7ex\hbox{E}\kern-.125emX}}

\begin{document}

\title{GeoCFNet: Geometry-Aware Confidence Field Network for Robot-Assisted Endoscopic Submucosal Dissection}

\author{Rui Tang*, Guankun Wang*, Long Bai, Haochen Yin, Huxin Gao, Jiewen Lai, Jiazheng Wang,  \\
Hongliang Ren,
\emph{Senior Member, IEEE}
\thanks{This work was supported by Hong Kong Research Grants Council (14200425, R4020-22, R1007-24, C4026-21GF, 14206125, 14204524, 14216022), the National Natural Science Foundation of China (62403402), the Guangdong Basic and Applied Research Foundation (2025A1515011594), the National Natural Science Foundation of China (T252500134), the Ministry of Science and Technology of China (2025YFE0122500), and the Hong Kong Innovation and Technology Fund (MHP/185/24). (\textit{Corresponding author: Hongliang Ren.})}

\thanks{R. Tang, G. Wang, L. Bai, H. Yin, H. Gao, J. Lai, and H. Ren are with the Department of Electronic Engineering, The Chinese University of Hong Kong, Hong Kong SAR, China.}
\thanks{J. Wang is with the Theory Lab, Central Research Institute, 2012 Labs, Huawei Technologies Co. Ltd., Hong Kong SAR, China}
}

\maketitle
\vspace{-1.5mm}

\begin{abstract}
Advanced surgical robotics has made robot-assisted endoscopic submucosal dissection (ESD) a promising approach for the en-bloc resection of large lesions, with the potential to reduce recurrence and improve long-term outcomes. However, the technical complexity and risk of complications in ESD demand stable and precise visual guidance to maintain an accurate dissection corridor and a safe tissue margin. Dense confidence fields provide an effective representation for this purpose by describing both the preferred dissection region and its spatial transition to surrounding tissue. However, reliable confidence field estimation remains challenging in dynamic endoscopic scenes due to smoke, specular highlights, tissue deformation, weak texture, and the thin geometric structure of the target region.
To address these challenges, we formulate dissection guidance as a geometry-aware confidence field estimation problem and propose GeoCFNet, a geometry-aware confidence field network built on a pretrained DINOv3 backbone. GeoCFNet integrates a Token-Differentiated Fusion module to aggregate class-token context with dense patch representations, a SegFormer decoder for confidence regression, and Geometry-Aware Spatial Regularization (GASR) to preserve spatial coherence and local geometric transitions. Experimental results show that GeoCFNet achieves RMSE 0.0480, PSNR 27.1995, SSIM 0.3397, and CC 0.2466, indicating accurate and geometrically stable confidence field estimation for robot-assisted ESD guidance.
\end{abstract}

\begin{IEEEkeywords}
Robot-Assisted Endoscopic Submucosal Dissection, Geometry-Aware Confidence Field Network, DINOv3, Anisotropic Regularization.
\end{IEEEkeywords}

\section{Introduction}
Robot-assisted endoscopic submucosal dissection (ESD) aims to improve the stability and precision of complex procedures~\cite{cao2023aiendo,meng2022robotic,tang2025geo,wang2025endoarss}. A key requirement is to maintain an accurate dissection trajectory within a narrow endoscopic field under dynamic intraoperative conditions. Errors in localizing the dissection region or following the intended submucosal plane may lead to incomplete resection, bleeding, or perforation~\cite{wang2025copesd,yilmaz2023emr,kim2024porcine}. Reliable intraoperative visual guidance is therefore essential for safe and consistent robot-assisted dissection.

\begin{figure}[!t]
\centering
\includegraphics[width=0.92\columnwidth]{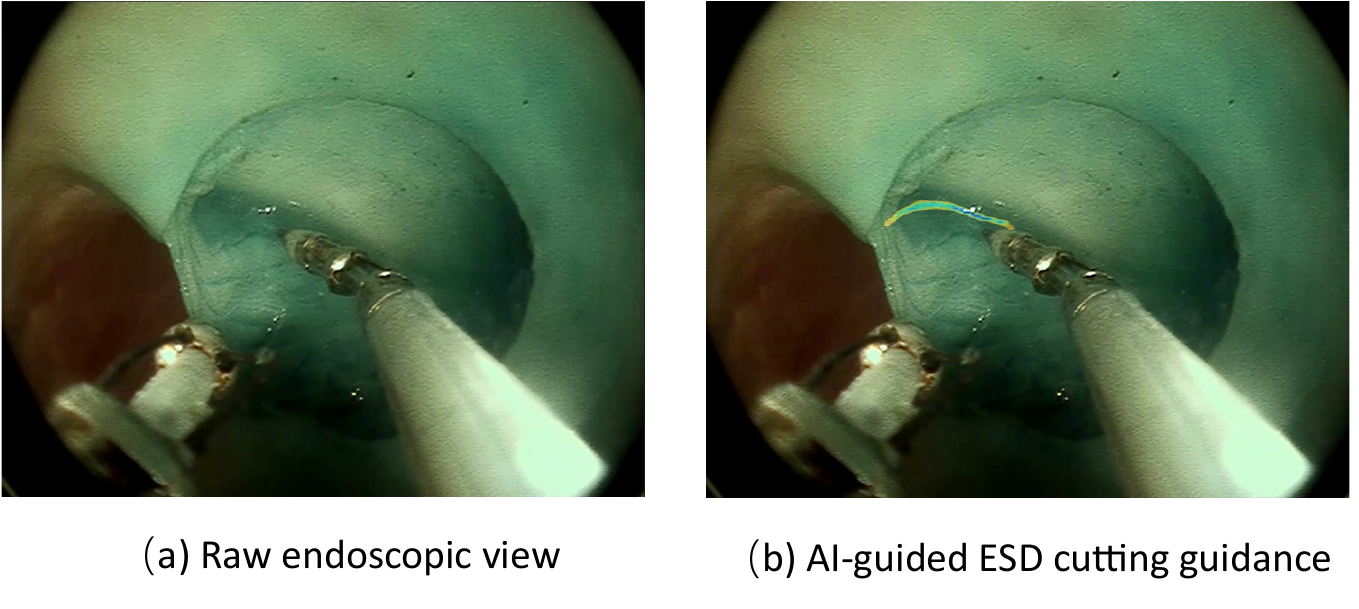}
\vspace{-2mm}
\caption{Illustration of AI-guided ESD cutting guidance. The raw endoscopic view contains ambiguous local visual cues, while the AI-guided overlay indicates the preferred cutting region for robot-assisted ESD.}
\vspace{-3mm}
\label{fig:intro_target}
\end{figure}

Existing visual guidance formulations are commonly based on point-level prediction~\cite{fathollahi2022video, li2023imitation} or binary segmentation~\cite{lin2022ds}, but these discrete representations provide limited information about the safety transition around the target region. Confidence field estimation addresses this limitation by encoding the preferred dissection region and its graded transition toward surrounding tissue, supporting safety-margin-aware intraoperative guidance~\cite{xu2025etsm}. As shown in Fig.~\ref{fig:intro_target}, the objective is to transform ambiguous endoscopic appearance into a structured guidance output with spatial continuity and geometric consistency.
This estimation remains challenging due to both visual ambiguity and geometric sensitivity. Smoke, specular reflection, motion blur, occlusion, bleeding, and tissue deformation reduce the reliability of local appearance cues~\cite{liu2019dense,cui2024surgical}. Meanwhile, the dissection target is often thin, elongated, and curved, making the confidence field sensitive to local regression errors. Such errors can cause discontinuous responses, boundary drift, or deformation of the intended guidance region~\cite{cui2024endodac,xu2025pdzseg}.

Prior work in surgical video analysis and endoscopic dense perception has advanced procedural understanding and pixel-level scene interpretation~\cite{blum2010modeling,padoy2012statistical}, yet geometry-consistent confidence estimation remains insufficiently explored. Workflow understanding and surgical phase recognition methods model procedural states from surgical videos~\cite{twinanda2016endonet,jin2017sv,czempiel2020tecno,gao2021trans}, but they do not generate dense guidance maps for intraoperative decision support. Encoder--decoder architectures such as U-Net and TransUNet are widely used for biomedical dense prediction~\cite{ronneberger2015unet,chen2024transunet}, but their objectives are usually defined for mask prediction rather than continuous confidence regression. Recent endoscopic perception methods introduce stronger visual priors and foundation-model adaptation for depth estimation and dissection-zone segmentation~\cite{cui2024surgical,cui2024endodac,xu2025pdzseg}. ETSM further studies confidence map-based safety-margin guidance for robot-assisted ESD~\cite{xu2025etsm}, but reliable confidence prediction under smoke, weak texture, and specular artifacts remains challenging. Existing methods still rarely couple transferable dense semantics with geometry-aware constraints designed for thin, elongated, and line-shaped confidence transitions.

Vision foundation models provide strong dense representations for addressing semantic ambiguity in endoscopic confidence estimation. DINOv2 demonstrates the transferability of large-scale self-supervised features for recognition and dense prediction~\cite{oquab2023dinov2}, and DINOv3 further improves dense feature quality and frozen-backbone transfer performance~\cite{simeoni2025dinov3}. Recent analysis of vision transformers also suggests that class-token information can complement patch-level descriptors in downstream prediction~\cite{marouani2026cls}. Nevertheless, pretrained semantics alone do not guarantee geometric coherence in confidence fields. With point-wise regression, predictions may still contain local noise, fragmented responses, or blurred transitions around the dissection corridor~\cite{simeoni2025dinov3,gupta2021efficient}. Therefore, reliable confidence estimation requires both transferable dense semantics and geometry-aware regularization.

To address these limitations, we define robot-assisted dissection guidance as geometry-aware confidence field estimation and propose GeoCFNet for dissection region confidence prediction. GeoCFNet integrates four components: a pretrained DINOv3 backbone for dense semantic encoding, a Token-Differentiated Fusion (TDF) module for global-to-local context aggregation, SegFormer decoding for spatial confidence reconstruction, and GASR for geometry-aware field regularization. The DINOv3 backbone provides transferable dense features for challenging endoscopic frames. TDF uses the class token as a global surgical-context prior to guide patch-level representations under ambiguous local appearance. The SegFormer decoder maps the enhanced features to a full-resolution confidence field. GASR further constrains the prediction by reducing local noise and preserving line-shaped transitions around the dissection region. The main contributions of this paper are summarized as follows:
\begin{itemize}
    \item We establish geometry-aware confidence field estimation as a dense guidance formulation for robot-assisted ESD, representing both the preferred dissection region and its safety-related spatial transition in a continuous field.
    \item We develop GeoCFNet by coupling DINOv3 dense representations with Token-Differentiated Fusion, enabling global surgical context to guide local confidence estimation under weak texture and visual artifacts.
    \item We design GASR with smoothness and anisotropic constraints, imposing structural priors that reduce local noise while preserving the line-shaped topology.
    \item Extensive experiments on robot-assisted ESD images demonstrate that GeoCFNet consistently outperforms representative baselines and produces spatially coherent confidence fields under visual artifacts, validating the effectiveness of the proposed framework.
\end{itemize}

\section{Method}
\noindent To address the challenges of unstable boundaries and noisy confidence maps, we propose GeoCFNet, a geometry-aware confidence field network for intraoperative confidence field estimation in robot-assisted ESD. The method uses a pretrained DINOv3 encoder, a Token-Differentiated Fusion (TDF) module, a confidence field decoder, and GASR to improve spatial stability. We first define the confidence field used for dissection guidance, and then present the overall architecture, the TDF module, and the GASR objective.

\subsection{Confidence Field Definition}
Given an intraoperative endoscopic image $\mathbf{I}\in\mathbb{R}^{H\times W\times 3}$, the target is a continuous confidence field $\mathbf{Y}\in[0,1]^{H\times W}$. Each value $Y_{ij}$ represents the relative confidence that pixel $(i,j)$ belongs to the preferred dissection guidance region. Higher values correspond to locations closer to the desired dissection region, whereas lower values indicate surrounding tissue with lower guidance preference. Compared with a binary mask, the confidence field preserves graded transitions around the target region and provides a continuous representation of the dissection region and its safety-related spatial margin.

The supervision target is constructed from the annotated dissection guidance region using nonlinear distance decay, where confidence values decrease progressively from the target region to surrounding tissue and are normalized to $[0,1]$. This formulation converts intraoperative guidance into a dense regression problem, requiring the model to estimate both the location of the preferred dissection region and the geometric continuity of its surrounding confidence transition.

\begin{figure*}[t]
\centering
\includegraphics[width=0.94\textwidth]{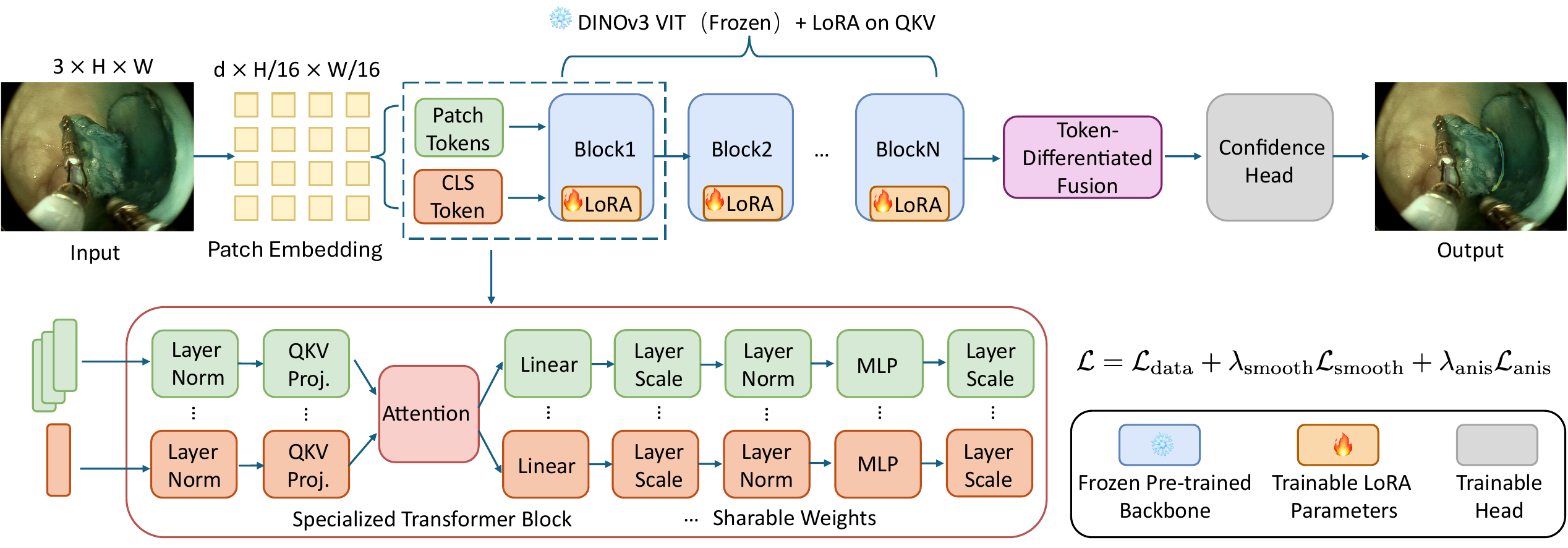}
\vspace{-2mm}
\caption{Overall architecture of GeoCFNet. A pretrained DINOv3 backbone extracts dense patch features and a CLS token from the input endoscopic image. The Token-Differentiated Fusion module uses the CLS token to enhance patch-level representations with global semantic context, and a SegFormer decoder reconstructs the full-resolution confidence field. During training, GASR imposes smoothness and anisotropic constraints to improve geometric coherence and preserve line-shaped transitions around the target dissection region.}
\vspace{-3mm}
\label{fig:method_overview}
\end{figure*}

\subsection{Overall Framework}
Existing pixel-wise regression methods mainly optimize numerical errors, but they do not explicitly model the spatial organization and geometric continuity of the confidence field. To address this limitation, Fig.~\ref{fig:method_overview} illustrates the overall architecture of GeoCFNet, which follows an encoder fusion decoder pipeline with GASR during training. The input image is tokenized into patch tokens and a CLS token by a pretrained DINOv3 encoder. The Token-Differentiated Fusion (TDF) module then fuses the class token with patch features to form an enhanced representation, and the confidence field decoder reconstructs the full-resolution confidence map. The pretrained DINOv3 backbone is frozen to preserve large-scale visual representations, while LoRA parameters in the QKV projections, the TDF module, and the confidence decoder are trainable for task adaptation.

Given an input image $\mathbf{I}\in\mathbb{R}^{H\times W\times 3}$, the DINOv3 encoder tokenizes the image into non-overlapping $16\times16$ patches and produces a sequence of visual tokens. We extract the hidden representation from the last transformer block,
\begin{equation}
\mathbf{T}^{(L)} = \mathcal{B}(\mathbf{I}),
\end{equation}
where $\mathbf{T}^{(L)}\in\mathbb{R}^{B\times (N+1)\times d}$, $B$ is the batch size, $N$ is the number of patch tokens, and $d$ is the feature dimension. The token sequence is decomposed into one class token and $N$ patch tokens:
\begin{equation}
\mathbf{T}^{(L)} = [\mathbf{c}^{(L)}, \mathbf{P}^{(L)}],
\end{equation}
where $\mathbf{c}^{(L)}\in\mathbb{R}^{B\times 1\times d}$ and $\mathbf{P}^{(L)}\in\mathbb{R}^{B\times N\times d}$. In our implementation, we use DINOv3-Base with $d=768$.

The patch tokens are reshaped into a two-dimensional feature map for dense prediction:
\begin{equation}
\mathbf{F}_{\mathrm{deep}}=
\mathrm{Reshape}(\mathbf{P}^{(L)})
\in\mathbb{R}^{B\times d\times H_p\times W_p},
\end{equation}
where $H_p=H/16$, $W_p=W/16$, and $N=H_pW_p$. This representation preserves the spatial layout of patch tokens and provides dense semantic features for confidence field estimation. The class token $\mathbf{c}^{(L)}$ is retained for TDF, where the class-token context is fused with the dense patch representation.

The TDF module produces an enhanced feature map $\tilde{\mathbf{F}}$, which is passed to a SegFormer decoder~\cite{xie2021segformer}. The decoder projects $\tilde{\mathbf{F}}$ into a lower-dimensional hidden space, refines the representation through feature fusion, and outputs a single-channel confidence map. The prediction is then bilinearly upsampled to the original image resolution:
\begin{equation}
\hat{\mathbf{Y}} = \mathrm{Up}\left(D(\tilde{\mathbf{F}})\right),
\end{equation}
where $D(\cdot)$ denotes the SegFormer decoder and $\mathrm{Up}(\cdot)$ denotes bilinear interpolation to resolution $H\times W$. The decoder reconstructs the confidence field from the semantically enhanced feature map, while the geometric regularization terms described below constrain the predicted field during training.

\subsection{Token-Differentiated Fusion}

In endoscopic scenes, visual degradations may obscure local evidence of the dissection region, making confidence prediction unstable when relying only on patch-level features. The target region is also constrained by global visual context, such as tissue layout, flap orientation, and the relative position of the dissection plane. The class token provides a global summary of the input image, and TDF uses learnable projections to adapt this generic visual context into task-specific guidance for interpreting corrupted local patches.

As illustrated in Fig.~\ref{fig:method_overview}, the DINOv3 transformer produces a class token and a set of patch tokens. The class token captures global visual information, whereas the patch tokens retain the spatial layout required for dense confidence estimation. Inspired by recent studies on class-patch token interaction in vision transformers~\cite{marouani2026cls}, TDF maps the class token into a task-adapted semantic embedding and uses it to refine the patch feature map for confidence field prediction.

Given the class token $\mathbf{c}^{(L)}$ from the last transformer block, TDF first maps it to a global semantic embedding:
\begin{equation}
\mathbf{s} = \phi(\mathbf{c}^{(L)}) \in \mathbb{R}^{B\times d},
\end{equation}
where $\phi(\cdot)$ denotes a learnable linear projection that adapts the class-token representation to the confidence estimation task. The embedding $\mathbf{s}$ is then broadcast to the spatial lattice of the patch feature map:
\begin{equation}
\mathbf{G} = \mathrm{Broadcast}(\mathbf{s}) \in \mathbb{R}^{B\times d\times H_p\times W_p}.
\end{equation}
The broadcast global context is adapted and fused with the dense patch representation:
\begin{equation}
\tilde{\mathbf{F}} = \mathbf{F}_{\mathrm{deep}} + \psi(\mathbf{G}),
\end{equation}
where $\psi(\cdot)$ denotes a learnable feature adaptation layer. Additive fusion incorporates class-token context without increasing the channel dimension or modifying the pretrained transformer, yielding an efficient dense representation aligned with the target confidence geometry.

\subsection{Geometry-Aware Spatial Regularization (GASR)}

The token fusion module improves semantic discrimination by incorporating image-level context into dense patch representations. However, semantic enhancement alone does not explicitly constrain the spatial structure of the predicted confidence field. When optimized only with point-wise supervision, the prediction may still contain local noise, fragmented responses, or blurred transitions around the target dissection region. To address this remaining limitation, we introduce GASR to impose spatial continuity while preserving line-shaped confidence transitions.

We use a composite objective that combines numerical supervision with geometric regularization:
\begin{equation}
\mathcal{L} = \mathcal{L}_{\mathrm{data}} + \lambda_{\mathrm{smooth}}\mathcal{L}_{\mathrm{smooth}} + \lambda_{\mathrm{anis}}\mathcal{L}_{\mathrm{anis}}.
\end{equation}

The data term is defined as the mean squared error between the predicted confidence field and the ground-truth map:
\begin{equation}
\mathcal{L}_{\mathrm{data}} = \frac{1}{HW}\sum_{i=1}^{H}\sum_{j=1}^{W}\left(\hat{Y}_{ij}-Y_{ij}\right)^2.
\end{equation}

To encourage spatial continuity, we impose a diffusion-style smoothness penalty on the gradients of the predicted field. Let $\partial_x$ and $\partial_y$ denote finite differences along the horizontal and vertical directions. We define
\begin{equation}
\mathcal{L}_{\mathrm{smooth}} =
\frac{1}{HW}\sum_{i,j}\left(
(\partial_x \hat{Y}_{ij})^2 + (\partial_y \hat{Y}_{ij})^2
\right).
\end{equation}
This term penalizes abrupt local variations and encourages smooth confidence changes in homogeneous regions.

Uniform smoothing may blur meaningful confidence transitions. We therefore introduce an anisotropic diffusion term guided by the ground-truth confidence gradient, whose nonlinear distance decay produces stronger variation near annotated transitions:
\begin{equation}
g^x_{ij}=\partial_x Y_{ij},\qquad g^y_{ij}=\partial_y Y_{ij}.
\end{equation}
The direction-dependent diffusion coefficients are defined as
\begin{equation}
\mathbf{c}^x_{ij}=\exp\left(-\frac{(g^x_{ij})^2}{\kappa_{\mathrm{anis}}^2}\right),\qquad
\mathbf{c}^y_{ij}=\exp\left(-\frac{(g^y_{ij})^2}{\kappa_{\mathrm{anis}}^2}\right),
\end{equation}
where $\kappa_{\mathrm{anis}}>0$ controls the sensitivity of the diffusion weights to structural variation. The anisotropic regularization term is formulated as
\begin{equation}
\mathcal{L}_{\mathrm{anis}} =
\frac{1}{HW}\sum_{i,j}\left(
\mathbf{c}^x_{ij}(\partial_x \hat{Y}_{ij})^2 +
\mathbf{c}^y_{ij}(\partial_y \hat{Y}_{ij})^2
\right).
\end{equation}

Large gradients near annotated transitions reduce the corresponding diffusion coefficients, preventing the loss from over-penalizing valid confidence changes. In contrast, low-gradient regions receive stronger regularization, where abrupt variations are more likely to be noise. The anisotropic coefficients are computed from the ground-truth confidence field only during training and are not required during inference. Together, $\mathcal{L}_{\mathrm{smooth}}$ and $\mathcal{L}_{\mathrm{anis}}$ suppress spurious fluctuations in homogeneous regions while allowing sharp, label-consistent transitions around the target dissection structure.

\begin{figure*}[t]
\centering
\includegraphics[width=0.88\textwidth]{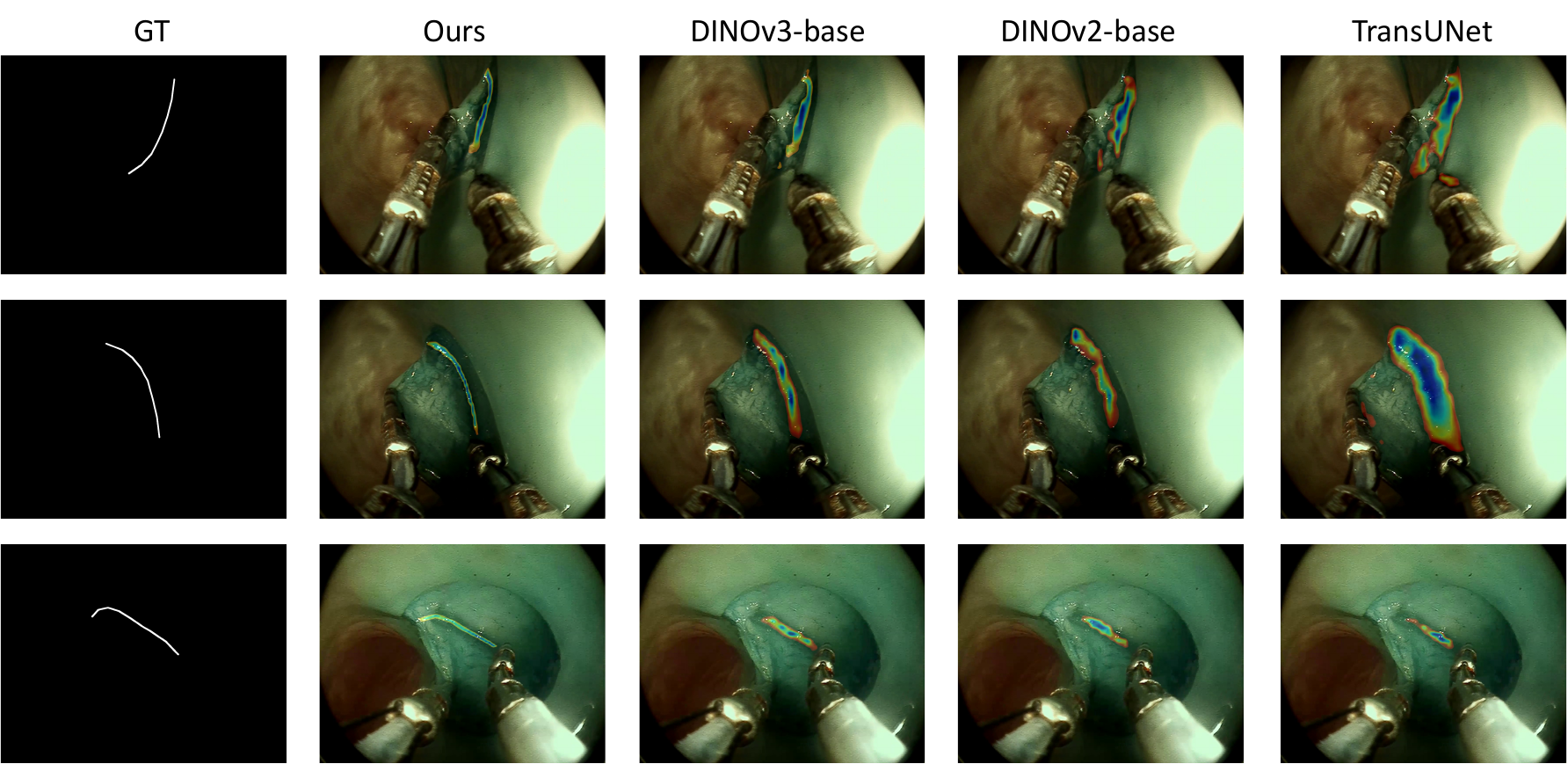}
\vspace{-2mm}
\caption{Qualitative comparison of confidence field estimation on representative endoscopic frames. The proposed method generates more spatially coherent confidence fields with reduced leakage to surrounding tissue and better alignment with the line-shaped target structure compared with DINOv3-base, DINOv2-base, and TransUNet.}
\vspace{-3mm}
\label{fig:qualitative_results}
\end{figure*}

\section{Experiments}
\subsection{Dataset}
The dataset contains 1{,}849 endoscopic images collected from robot-assisted ESD scenes~\cite{xu2025etsm}. We split the dataset into 1{,}480 training images and 369 test images. Each image is paired with a ground-truth confidence map that represents the target dissection region and its surrounding confidence transition. The same training and test split is used for all compared methods to ensure a fair evaluation.
\subsection{Implementation Details}
All experiments are conducted on a single NVIDIA RTX 4090 GPU. Each model is trained for 100 epochs using the Adam optimizer with a cosine annealing schedule. The backbone of GeoCFNet is initialized with pretrained DINOv3 weights. LoRA is inserted into the QKV projections of the DINOv3 transformer blocks with rank \(r=4\). The pretrained backbone is frozen, and only the LoRA parameters, TDF module, and confidence decoder are updated during training. All competing methods are trained under the same data split and basic training protocol. We report RMSE~\cite{chai2014rmse}, PSNR~\cite{huynh2008scope}, SSIM~\cite{wang2004image}, and CC~\cite{pearson1895note} to evaluate pixel-level accuracy, reconstruction quality, local geometric consistency, and global distribution agreement, respectively.

\subsection{Results}
\subsubsection{Main Comparison}
We compare GeoCFNet with representative dense prediction baselines, including U-Net~\cite{ronneberger2015unet}, TransUNet~\cite{chen2024transunet}, DINOv2-small/base~\cite{oquab2023dinov2}, and DINOv3-small/base~\cite{simeoni2025dinov3}. GeoCFNet uses DINOv3-base with the proposed TDF module, a confidence decoder, and GASR.
Table~\ref{tab:main_results} reports the quantitative comparison. GeoCFNet achieves the best performance across all four metrics. Compared with U-Net and TransUNet, the foundation-model baselines obtain lower RMSE and higher structural metrics, showing the benefit of pretrained dense representations under weak texture and large appearance variation. DINOv3-small and DINOv3-base further outperform DINOv2-small and DINOv2-base, suggesting stronger transferability for dissection-region confidence estimation. Compared with the strongest DINOv3 baseline, GeoCFNet reduces RMSE from 0.0801 to 0.0480 and improves PSNR from 22.3018 to 27.1995, while also achieving additional gains in SSIM and CC.

Fig.~\ref{fig:qualitative_results} presents qualitative comparisons on representative endoscopic frames. Compared with DINOv2-base and TransUNet, which generate diffuse responses and leakage into surrounding tissue, GeoCFNet produces more concentrated confidence fields around the target dissection region. Compared with DINOv3-base, GeoCFNet further improves alignment with the annotated line-shaped guidance structure.

\begin{table}[t]
\centering
\caption{Comparison of confidence field estimation performance across different methods.}
\begin{tabular}{lcccc}
\toprule
\textbf{Method} & \textbf{RMSE$\downarrow$} & \textbf{PSNR$\uparrow$} & \textbf{SSIM$\uparrow$} & \textbf{CC$\uparrow$} \\
\midrule
U-Net~\cite{ronneberger2015unet}        & 0.1510 & 16.0430 & 0.2604 & 0.2109 \\
TransUNet~\cite{chen2024transunet}    & 0.1009 & 19.0649 & 0.2793 & 0.2210 \\
DINOv2-small~\cite{oquab2023dinov2} & 0.1257 & 18.3119 & 0.2749 & 0.1873 \\
DINOv2-base~\cite{oquab2023dinov2}  & 0.1250 & 18.8553 & 0.2880 & 0.2198 \\
DINOv3-small~\cite{simeoni2025dinov3} & 0.0901 & 21.1935 & 0.3212 & 0.2203 \\
DINOv3-base~\cite{simeoni2025dinov3}  & 0.0801 & 22.3018 & 0.3196 & 0.2125 \\
GeoCFNet    & \textbf{0.0480} & \textbf{27.1995} & \textbf{0.3397} & \textbf{0.2466} \\
\bottomrule
\end{tabular}
\label{tab:main_results}
\end{table}

\subsubsection{Ablation Studies}
Table~\ref{tab:ablation_core} evaluates the effects of the TDF module and the proposed GASR. Starting from the DINOv3 baseline, adding TDF reduces RMSE from 0.0801 to 0.0501 and improves PSNR from 22.3018 to 26.8803, showing that class-token context aggregation substantially improves dense confidence estimation under ambiguous local appearance. The GASR-only variant also improves over the baseline, with RMSE decreasing to 0.0542, indicating that GASR contributes to more stable confidence prediction.
Combining TDF with GASR achieves the best overall performance, with RMSE 0.0480, PSNR 27.1995, SSIM 0.3397, and CC 0.2466. Compared with DINOv3+TDF, the full model further improves SSIM and CC, suggesting better structural consistency and global correlation. These results indicate that TDF and GASR provide complementary benefits: TDF enhances semantic disambiguation, while the regularization terms improve the spatial coherence of the predicted confidence field.
\begin{table}[t]
\centering
\caption{Ablation study of the core components in GeoCFNet.}
\begin{tabular}{lcccc}
\toprule
\textbf{Method} & \textbf{RMSE$\downarrow$} & \textbf{PSNR$\uparrow$} & \textbf{SSIM$\uparrow$} & \textbf{CC$\uparrow$} \\
\midrule
DINOv3       & 0.0801 & 22.3018 & 0.3196 & 0.2125 \\
DINOv3 + TDF  & 0.0501 & 26.8803 & 0.3251 & 0.2312 \\
DINOv3 + GASR & 0.0542 & 24.3731 & 0.3205 & 0.2237 \\
DINOv3 + TDF + GASR & \textbf{0.0480} & \textbf{27.1995} & \textbf{0.3397} & \textbf{0.2466} \\
\bottomrule
\end{tabular}
\label{tab:ablation_core}
\end{table}

\subsubsection{Fine-tuning Experiments on Loss Sensitivity}
We analyze the sensitivity of anisotropic regularization to $\lambda_{\mathrm{anis}}$ and $\kappa_{\mathrm{anis}}$ in Table~\ref{tab:loss_sensitivity}. For $\lambda_{\mathrm{anis}}$, removing the anisotropic constraint limits structural consistency, while $\lambda_{\mathrm{anis}}=0.1$ gives the best overall result; larger weights provide no further gain and may over-constrain local confidence regression.
For $\kappa_{\mathrm{anis}}$, the best result is obtained at $\kappa_{\mathrm{anis}}=0.2$. Smaller values make the diffusion weights overly sensitive to local ground-truth gradients, whereas larger values make the anisotropic term closer to uniform smoothing. This trend shows that gradient-guided diffusion should preserve line-shaped confidence transitions without suppressing valid local variations.
\begin{table}[t]
\centering
\caption{Sensitivity analysis of anisotropic regularization parameters.}
\begin{tabular}{lcccc}
\toprule
\textbf{$\lambda_{\mathrm{anis}}$} & \textbf{RMSE$\downarrow$} & \textbf{PSNR$\uparrow$} & \textbf{SSIM$\uparrow$} & \textbf{CC$\uparrow$} \\
\midrule
0   & 0.0627 & 25.0611 & 0.3233 & 0.2311 \\
0.1 & \textbf{0.0480} & \textbf{27.1995} & \textbf{0.3397} & \textbf{0.2466} \\
0.2 & 0.0507 & 26.0935 & 0.3353 & 0.2351 \\
0.5 & 0.0505 & 26.1216 & 0.3297 & 0.2395 \\
\midrule
\textbf{$\kappa_{\mathrm{anis}}$} & \textbf{RMSE$\downarrow$} & \textbf{PSNR$\uparrow$} & \textbf{SSIM$\uparrow$} & \textbf{CC$\uparrow$} \\
\midrule
0   & 0.0511 & 26.7386 & 0.3128 & 0.2255 \\
0.1 & 0.0510 & 27.0189 & 0.3191 & 0.2276 \\
0.2 & \textbf{0.0480} & \textbf{27.1995} & \textbf{0.3397} & \textbf{0.2466} \\
0.5 & 0.0509 & 26.0609 & 0.3179 & 0.2374 \\
\bottomrule
\end{tabular}
\label{tab:loss_sensitivity}
\end{table}

\section{Conclusion}
In this work, we introduced GeoCFNet, a geometry-aware confidence field network for robot-assisted ESD. By integrating pretrained DINOv3 dense representations, Token-Differentiated Fusion, and GASR, GeoCFNet addresses visual ambiguity and geometric instability in endoscopic confidence prediction. The proposed framework improves both numerical accuracy and spatial coherence, producing confidence fields that better preserve line-shaped transitions around the target dissection region. Extensive experiments demonstrate the effectiveness of combining foundation-model semantics with geometry-aware regularization for reliable dissection guidance. This work highlights the potential of geometry-aware confidence estimation for safer intraoperative decision-making and provides a foundation for developing more robust robot-assisted endoscopic guidance systems.

\begingroup
\scriptsize
\bibliographystyle{IEEEtran}
\bibliography{reference}
\endgroup

\end{document}